\documentclass[letterpaper]{article} 
\usepackage{aaai2027}  
\usepackage[hyphens]{url}  
\usepackage{graphicx} 
\usepackage{natbib}  
\usepackage{caption} 
\usepackage{algorithm}
\usepackage{algorithmic}
\usepackage{booktabs}
\usepackage{multirow}
\usepackage{graphicx}
\usepackage{amsmath}
\usepackage{amssymb}
\newcommand{\abs}[1]{\left\lvert #1 \right\rvert}
\newcommand{\norm}[1]{\left\lVert #1 \right\rVert}
\usepackage{newfloat}
\usepackage{listings}
\DeclareCaptionStyle{ruled}{labelfont=normalfont,labelsep=colon,strut=off} 
\floatstyle{ruled}
\newfloat{listing}{tb}{lst}{}
\floatname{listing}{Listing}

\usepackage{booktabs}

\title{DuetHOI: Language-Guided Bimanual Hand--Object Motion Generation with Articulation Planning and Contact Refinement}
\author{
    Zhi Wang\textsuperscript{\rm 1},
    Yuyan Liu\textsuperscript{\rm 1},
    Liu Liu\textsuperscript{\rm 1}\corresponding,
    Ruonan Liu\textsuperscript{\rm 2},
    Ruixuan Liu\textsuperscript{\rm 3}
}

\affiliations{
    \textsuperscript{\rm 1}Hefei University of Technology\\
    \textsuperscript{\rm 2}Shanghai Jiao Tong University\\
    \textsuperscript{\rm 3}Anhui University\\
    liuliu@hfut.edu.cn
}

\begin{document}

\maketitle

\begin{figure*}[tb]
    \centering
    \includegraphics[width=\textwidth]{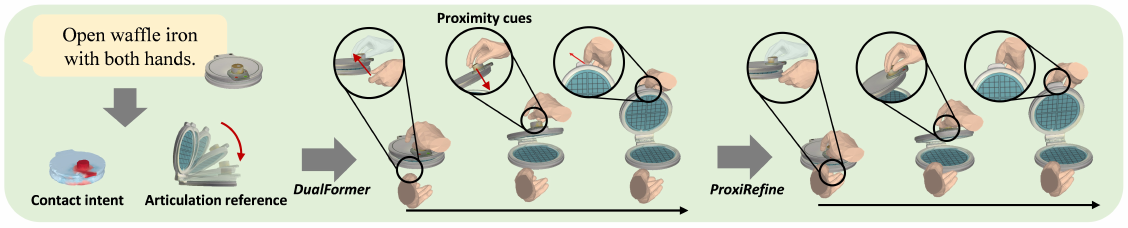}
    \caption{DuetHOI generation process.
    Contact intent and an articulation reference guide global interaction generation, after which ProxiRefine keeps the object trajectory fixed while correcting local bimanual geometry with proximity cues.}
    \label{fig:teaser}
\end{figure*}

\begin{abstract}
Bimanual articulated-object interaction generation requires a model to capture the evolution of object articulation, coordination between the two hands, and fine-grained hand--object contact. Existing methods typically encode object and hand motion as a unified high-dimensional sequence, making it difficult to explicitly accommodate the different scales of manipulation progress, relative bimanual motion, and local hand pose. We propose \textbf{DuetHOI}, a structured framework for bimanual articulated-object interaction generation. DuetHOI first uses ContactVAE to predict contact intent on the object surface from language and object geometry. Conditioned on this spatial intent, ArtPlanner provides a trajectory-level articulation reference, and DualFormer generates the global interaction using object states, object-centered hand positions, and compact hand-pose tokens learned by ManiVAE. After global generation, ProxiRefine fixes the object trajectory and residually corrects the two hands using current hand--object proximity, improving local surface alignment. Experiments across articulated and rigid objects and single- and bimanual settings show that DuetHOI outperforms three adapted baselines on most contact and hand--object consistency metrics, with particularly strong results in bimanual articulated interaction generation.
\end{abstract}

\section{Introduction}
\label{sec:intro}

Opening a laptop, operating a microwave, or rotating a movable object part requires two hands to coordinate around a changing object state. Such language-driven motion generation is important for dexterous robotics \cite{zhang2024artigrasp,christen2022d,li2025maniptrans}, embodied agents \cite{li2025object}, and virtual character animation \cite{zhang2025mikudance}. Language-driven bimanual articulated-object interaction generation is the focus of this work: given an object and a language instruction, the model infers action regions on the object surface and generates coherent object and two-hand motion.

Three coupled challenges arise in bimanual articulated manipulation: \textbf{articulation progress}, for which object-state errors alter the subsequent manipulation; \textbf{bimanual coordination}, which assigns complementary hand roles in the object-centered frame; and \textbf{local contact geometry}, where finger-pose errors cause contact failures. Recent data-driven HOI methods commonly concatenate object motion, hand trajectories, and hand poses into a unified sequence \cite{jiang2021hand,li2024semgrasp,cha2024text2hoi,li2025latenthoi,ye2023affordance,huang2025hoigpt}. This representation entangles large-scale state changes with fine-grained hand motion and offers limited control for correcting local contact after the global object trajectory is generated \cite{grady2021contactopt,jiang2021hand,liu2023contactgen}. In addition, an instruction can admit several valid contact locations, hand-role assignments, and trajectories \cite{ye2023affordance,yang2024learning,liu2023contactgen}.

These challenges motivate decomposing the interaction at both the representation and generation levels. Object state, object-centered hand positions, and local hand poses describe manipulation progress, bimanual coordination, and contact configuration, respectively. The model first generates the global interaction and then refines local hand geometry with the generated object trajectory held fixed, preserving manipulation progress and improving hand--object alignment.

Accordingly, we propose \textbf{DuetHOI}. ContactVAE first infers object-side contact intent from language and canonical object geometry and represents it as a spatial contact map. Conditioned on this spatial prior, language, object geometry, and the initial articulation state, ArtPlanner predicts a trajectory-level articulation reference. ManiVAE encodes high-dimensional hand poses into compact latent tokens. DualFormer organizes object state, object-centered left/right hand positions, and left/right hand-pose tokens as a token--time grid, and generates the global interaction conditioned on the contact map and articulation reference. The generated object trajectory is then held fixed while ProxiRefine predicts hand residuals from current hand--object proximity to correct local surface alignment.

Experiments evaluate DuetHOI on articulated and rigid objects in single- and bimanual settings. The results show consistent improvements in contact quality, hand accuracy, and relative hand--object consistency, with particularly strong performance in bimanual articulated-object interactions.

Our main contributions are:
\begin{itemize}
    \item We propose DuetHOI, a structured framework that represents object state, object-centered bimanual positions, and compact hand poses in separate streams, and stages global interaction generation with object-fixed local hand refinement.
    \item We introduce ContactVAE and ArtPlanner to provide complementary spatial and temporal conditions: a predicted object-side contact map and a contact-conditioned articulation reference.
    \item We develop ManiVAE, DualFormer, and ProxiRefine for structured generation and refinement: latent pose tokens and token--time attention model bimanual interaction, while proximity-guided residuals refine local hand--object alignment without altering object motion.
\end{itemize}

\section{Related Work}
\label{sec:related_work}

\subsection{Bimanual Hand Object Interaction}
\label{sec:related_bimanual_hoi}

HOI generation methods are broadly physics-based or data-driven. Physics-based methods learn manipulation policies through reinforcement learning in simulation \cite{zhang2024graspxl,zhang2024artigrasp,xu2023unidexgrasp,christen2022d}, but depend on simulator fidelity and reward design, which can limit generalization. Data-driven methods learn HOI distributions from multimodal motion data \cite{yang2022oakink,taheri2022goal,liu2021synthesizing,corona2020ganhand,turpin2022grasp,yang2024learning,zheng2023cams}, commonly using autoencoder-based \cite{li2024semgrasp,huang2025hoigpt,jiang2021hand,taheri2020grab,wu2022saga} or diffusion-based models \cite{li2025latenthoi,cha2024text2hoi}. Most jointly predict object motion, hand trajectories, articulation, and hand poses in a single representation, obscuring the distinct roles of global manipulation progress, bimanual coordination, and local contact. DuetHOI separates articulation planning, latent bimanual generation, and hand-pose refinement.

\subsection{Human Motion Priors and Pose Refinement}
\label{sec:related_motion_priors_refinement}

Learned motion priors reduce the complexity of high-dimensional pose generation through variational, discrete, or diffusion-based latent representations \cite{ludwig2025efficient,rempe2021humor,petrovich2021action,petrovich2022temos,guo2022generating,tevet2022motionclip,zhou2024emdm,zhang2023generating,guo2024momask}. While these priors mainly regularize body motion, bimanual articulated HOI also demands precise contact and coordination with object articulation. Refinement methods correct local pose errors either generically \cite{moon2019posefix} or with contact prediction, test-time adaptation, and differentiable optimization for hand--object interaction \cite{grady2021contactopt,jiang2021hand,liu2023contactgen}. DuetHOI combines both ideas: ManiVAE encodes compact hand-pose tokens, and ProxiRefine corrects the generated hands in an object-fixed frame while preserving object motion.

\begin{figure*}[tb]
    \centering
    \includegraphics[width=0.98\textwidth]{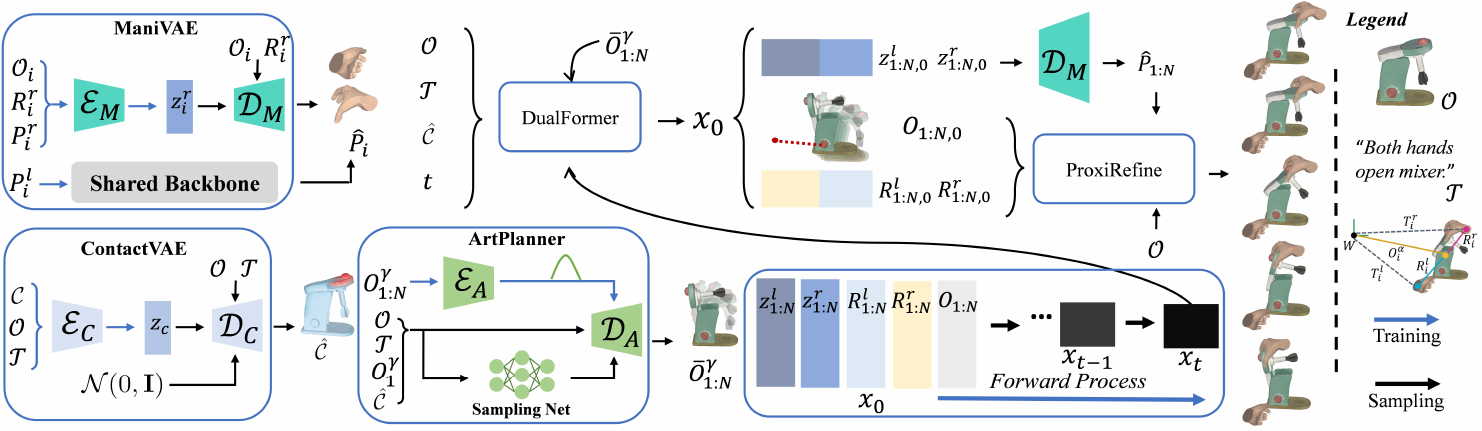}
    \caption{DuetHOI pipeline.
    ContactVAE predicts an object-side contact map that conditions both ArtPlanner and DualFormer. ArtPlanner provides a trajectory-level articulation reference for structured DualFormer generation, ManiVAE maps between hand poses and compact tokens, and ProxiRefine residually corrects the generated hands while preserving object motion. The lower-right inset visualizes the object-centered hand-position representation, with $W$ denoting the world origin.}
    \label{fig:pipeline}
\end{figure*}

\section{Method}
\subsection{Overview of DuetHOI}
\label{sec:method_overview}

Given object geometry $\mathcal{O}$ and a language instruction $\mathcal{T}$, DuetHOI first predicts an object-side contact map $\hat{\mathcal C}\in[0,1]^M$ and uses it to condition articulation planning and global interaction generation. The resulting output is an $N$-frame sequence of articulated object states and bimanual motion.
At frame $i$, the hand state $H_i=\{T_i,P_i\}$ contains global translations $T_i=\{T_i^l,T_i^r\}$ and MANO poses \cite{romero2022embodied} $P_i=\{P_i^l,P_i^r\}$. The object state $O_i=\{O_i^\alpha,O_i^\beta,O_i^\gamma\}$ contains a 3D translation $O_i^\alpha\in\mathbb{R}^3$, a rot6d rotation \cite{zhou2019continuity} $O_i^\beta\in\mathbb{R}^6$, and a scalar articulation coordinate $O_i^\gamma\in\mathbb{R}$, which is set to zero for rigid objects. To remove the world translation shared by the object and hands, we use object-centered hand positions $R_i^h=T_i^h-O_i^\alpha$ for $h\in\{l,r\}$.
Object geometry features are constructed in the same reference frame, allowing them and the relative hand positions to directly characterize hand--object spatial relations. This representation removes shared translation while preserving world-frame orientation; global hand translations are recovered by $\hat{T}_i^h=\hat{O}_i^\alpha+\hat{R}_i^h$.
Text and geometry are encoded by frozen CLIP \cite{radford2021learning} and PointNet++ \cite{qi2017pointnet++}, respectively, except that ContactVAE uses a point-wise MLP.

Figure~\ref{fig:pipeline} shows the full pipeline. ContactVAE predicts an object-side contact map that captures contact intent. ArtPlanner uses this spatial prior to provide an articulation reference, while ManiVAE represents hand articulation with compact tokens. Conditioned on the contact map and articulation reference, DualFormer generates the global interaction, after which ProxiRefine adjusts the hands with the object trajectory fixed.

\subsection{Language-Conditioned Contact Map Prediction}
\label{sec:contactvae}

Language specifies manipulation intent but not the precise object regions acted on by the hands \cite{ye2023affordance,yang2024learning,liu2023contactgen}. We sample $M$ object-surface points and define a static spatial contact map $\mathcal C\in\{0,1\}^M$ that represents potential contact regions throughout the interaction. Here, $\mathcal C_m$ indicates whether the $m$th point belongs to such a region. ContactVAE predicts the corresponding probabilities $\hat{\mathcal C}\in[0,1]^M$ from the instruction and object point cloud. Its encoder infers $q_{\phi_C}(z_C\mid\mathcal C,\mathcal O,\mathcal T)$ during training, and the decoder reconstructs $\mathcal C$. Because contact is sparse, we combine binary cross-entropy with Dice reconstruction \cite{milletari2016v}:
{\small
\begin{equation}
\begin{aligned}
\mathcal L_{\mathrm{BCE}}
&=-\frac{1}{M}\sum_{m=1}^{M}\left[
\mathcal C_m\log\hat{\mathcal C}_m
+(1-\mathcal C_m)\log(1-\hat{\mathcal C}_m)\right],\\
\mathcal L_{\mathrm{Dice}}
&=1-\frac{2\sum_m\hat{\mathcal C}_m\mathcal C_m+\varepsilon_C}
{\sum_m\hat{\mathcal C}_m+\sum_m\mathcal C_m+\varepsilon_C}.
\end{aligned}
\end{equation}
}
The constant $\varepsilon_C>0$ stabilizes the Dice ratio.
The latent posterior is regularized for inference-time sampling by
{\small
\begin{equation}
\mathcal L_C^{\mathrm{KL}}
=D_{\mathrm{KL}}\!\left(
q_{\phi_C}(z_C\mid\mathcal C,\mathcal O,\mathcal T)
\,\|\,\mathcal N(0,\mathbf I)\right).
\end{equation}
}
At inference, the contact map is generated as
{\small
\begin{equation}
z_C\sim\mathcal N(0,\mathbf I),\qquad
\hat{\mathcal C}=D_C(z_C;\mathcal O,\mathcal T).
\end{equation}
}
The principal ContactVAE objective is
{\small
\begin{equation}
\mathcal L_C
=\mathcal L_{\mathrm{BCE}}
+\mathcal L_{\mathrm{Dice}}
+\beta_C\mathcal L_C^{\mathrm{KL}}.
\end{equation}
}
The predicted contact probabilities condition both ArtPlanner and DualFormer.

\subsection{Articulation Trajectory Planning}
\label{sec:articulation_prior}

Object articulation provides a low-dimensional description of manipulation progress. ArtPlanner predicts its scalar trajectory $O^\gamma_{1:N}$ before full hand--object generation, conditioned on language $\mathcal T$, object geometry $\mathcal O$, the initial articulation $O_1^\gamma$, and the predicted contact probabilities. The scalar represents an angle for revolute joints and a displacement for prismatic joints; rigid objects use $O^\gamma_{1:N}=\mathbf 0$.

Let $c_A=(\mathcal{O},\mathcal{T},O_1^\gamma,\hat{\mathcal C})$. During training, the encoder maps the ground-truth trajectory and condition to
{\small
\begin{equation}
q_{\phi_A}(z_A\mid O^\gamma_{1:N},c_A)=\mathcal{N}(\mu_A^q,\sigma_A^q),
\end{equation}
}
and reconstructs the training trajectory from a posterior sample:
{\small
\begin{equation}
z_A^q\sim q_{\phi_A}(z_A\mid O^\gamma_{1:N},c_A),\qquad
\bar{O}^{\gamma}_{1:N}=D_A(z_A^q;c_A).
\end{equation}
}
The posterior-decoded reference is supervised by
{\small
\begin{equation}
\mathcal{L}_{\mathrm{traj}} =
\frac{1}{N}\sum_{i=1}^{N}
\abs{\bar{O}^{\gamma}_i-O^{\gamma}_i}.
\end{equation}
}
To generate an articulation reference from conditions alone at test time, we also learn a conditional prior
{\small
\begin{equation}
p_{\psi_A}(z_A\mid c_A)=\mathcal{N}(\mu_A^p,\sigma_A^p).
\end{equation}
}
Here $(\mu_A^q,\sigma_A^q)$ and $(\mu_A^p,\sigma_A^p)$ parameterize the posterior and conditional-prior Gaussians, respectively.
The conditional latent space is aligned by
{\small
\begin{equation}
\mathcal L_A^{\mathrm{KL}}
=D_{\mathrm{KL}}\!\left(q_{\phi_A}(z_A\mid O^\gamma_{1:N},c_A)\,\|\,p_{\psi_A}(z_A\mid c_A)\right).
\end{equation}
}
The principal ArtPlanner objective is
{\small
\begin{equation}
\mathcal L_A
=\mathcal L_{\mathrm{traj}}
+\beta_A\mathcal L_A^{\mathrm{KL}}.
\end{equation}
}
At inference, the deterministic main pipeline decodes the conditional-prior mean:
{\small
\begin{equation}
\bar{O}^{\gamma}_{1:N}=D_A(\mu_A^p;c_A).
\end{equation}
}
For stochastic inference, we sample $z_A^p\sim p_{\psi_A}(z_A\mid c_A)$ and decode $\bar O^\gamma_{1:N}=D_A(z_A^p;c_A)$ as the articulation reference; rigid interactions use $\bar O^\gamma_{1:N}=\mathbf 0$. DualFormer produces the final articulation prediction.

\subsection{Local Hand Pose Prior Learning}
\label{sec:hand_pose_prior}

Directly generating frame-wise MANO rotations substantially enlarges the structured motion space. ManiVAE encodes local hand articulation into a compact token \cite{li2024semgrasp,li2025latenthoi}. We denote by $\mathcal O_i$ the object geometry $\mathcal O$ transformed by the rotation $O_i^\beta$ and articulation $O_i^\gamma$ at frame $i$, with the object translation $O_i^\alpha$ set to zero. Both the encoder and decoder are conditioned on the object-centered hand position $R_i^h$ and the transformed object geometry $\mathcal O_i$. These conditions allow the latent token to focus on local hand articulation without redundantly encoding global placement, yielding a more compact pose representation and encouraging the decoded articulation to remain consistent with the current hand--object configuration.

We train one ManiVAE for both hands in a canonical right-hand space. Right-hand samples remain unchanged; left-hand samples are mirrored before encoding and mapped back after decoding. Given the canonicalized pose $P_i^h$, ManiVAE encodes a hand-pose token and reconstructs the MANO articulation:
{\small
\begin{equation}
\begin{aligned}
z_i^h&\sim q_{\phi_M}(z_i^h\mid P_i^h,R_i^h,\mathcal O_i),\\
\hat P_i^h&=D_M(z_i^h,R_i^h,\mathcal O_i).
\end{aligned}
\end{equation}
}
Global hand translation is recovered from the generated object translation and object-centered hand position.

We train ManiVAE with parameter- and geometry-space reconstruction. The former directly supervises the 6D MANO representation,
{\small
\begin{equation}
\mathcal{L}_{\mathrm{param}}^M =
\norm{P_i^h-\hat{P}_i^h}_2^2.
\end{equation}
}
Let $\mathcal V(\cdot)$ and $\mathcal J(\cdot)$ denote MANO vertices and joints. We supervise the decoded hand geometry by
{\small
\begin{equation}
\mathcal L_{\mathrm{geom}}^M
=\norm{\mathcal V(\hat P_i^h)-\mathcal V(P_i^h)}_2^2
+\norm{\mathcal J(\hat P_i^h)-\mathcal J(P_i^h)}_2^2.
\end{equation}
}
The latent space is regularized by
{\small
\begin{equation}
\mathcal L_M^{\mathrm{KL}} =
D_{\mathrm{KL}}
\left(
q_{\phi_M}(z_i^h\mid P_i^h,R_i^h,\mathcal O_i)
\,\|\,\mathcal{N}(0,\mathbf{I})
\right).
\end{equation}
}
The compact objective is
{\small
\begin{equation}
\mathcal{L}_M
=\mathcal L_{\mathrm{param}}^M
+\mathcal L_{\mathrm{geom}}^M
+\beta_M\mathcal L_M^{\mathrm{KL}}.
\end{equation}
}

We freeze ManiVAE and use the posterior mean $z_i^h=\mu_{\phi_M}(P_i^h,R_i^h,\mathcal O_i)$ as the hand-token target for DualFormer.

\begin{figure}[tb]
    \centering
    \includegraphics[width=\linewidth]{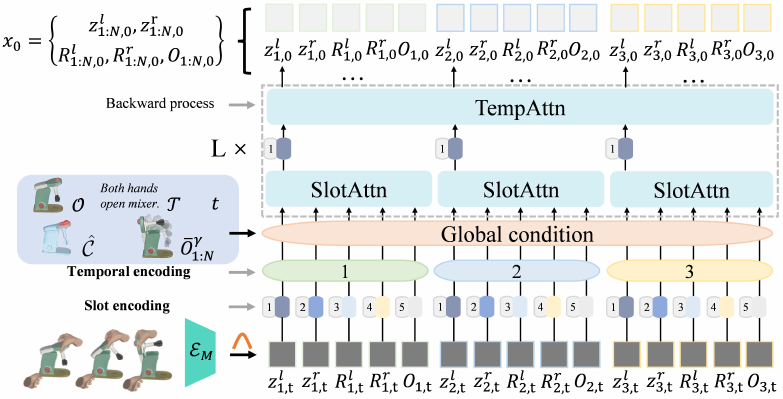}
    \caption{DualFormer.
    Slot attention models within-frame hand--object coordination across the five semantic streams, while temporal attention captures the evolution of each stream throughout the denoising process.}
    \label{fig:dualformer}
\end{figure}

\subsection{Token-Temporal Interaction Generation}
\label{sec:token_temporal_generation}

DualFormer generates the global interaction sequence on a token-temporal grid. After ArtPlanner provides an articulation reference and ManiVAE maps hand articulation into compact latent tokens, the clean diffusion target is represented with five semantic slots:
{\small
\begin{equation}
x_0 =
\left[
z^l_{1:N,0},\,
z^r_{1:N,0},\,
R^l_{1:N,0},\,
R^r_{1:N,0},\,
O_{1:N,0}
\right].
\end{equation}
}
The five streams separate object motion, relative hand placement, and local hand articulation. We standardize each stream with training-set statistics and apply stream-specific input and output projections. The DDPM \cite{ho2020denoising} forward process gives
{\small
\begin{equation}
x_t =
\sqrt{\bar{\alpha}_t}x_0
+ \sqrt{1-\bar{\alpha}_t}\epsilon,
\qquad
\epsilon\sim\mathcal{N}(0,\mathbf{I}),
\end{equation}
}
where $t$ is the diffusion timestep and $\bar{\alpha}_t$ is the cumulative noise schedule.

A naive flattened-attention design jointly models all $NS$ tokens. DualFormer instead factorizes this operation into separate attention operations for within-frame coordination and temporal propagation. Each component $x_{i,t}^{a}$ at frame $i$ and slot $a\in\{1,\ldots,S\}$ is mapped into a shared hidden space by the slot-specific projection $e_{i,t}^{a}=\mathrm{Proj}_{a}(x_{i,t}^{a})+\mathrm{PE}_{\mathrm{time}}(i)$, which also encodes its stream identity.
At a fixed diffusion timestep $t$, the projected streams form $e_t\in\mathbb{R}^{N\times S\times d}$, where $S=5$ is the number of semantic streams and $d$ is the hidden dimension. Inspired by factorized video modeling~\cite{arnab2021vivit}, at each frame $i$, slot attention jointly processes the $S$ tokens $(e_{i,t}^{1},\ldots,e_{i,t}^{S})$ to model hand--object coordination. For each semantic slot $a$, temporal attention then processes $(e_{1,t}^{a},\ldots,e_{N,t}^{a})$ across the $N$ frames. Each DualFormer block applies these two operations sequentially. Both operations are modulated by the DualFormer condition $c_D$. This factorization reduces the attention cost from $O(N^2S^2)$ for flattened attention to $O(NS^2+SN^2)$.

Let $c_D=(\mathcal O,\mathcal T,\hat{\mathcal C},\bar O^\gamma_{1:N},t)$ denote the DualFormer condition.
It combines object geometry, language, predicted contact probabilities, the ArtPlanner trajectory, and the diffusion timestep. Each value in $\hat{\mathcal C}$ is appended to its corresponding object point before encoding. Adaptive normalization injects $c_D$ into each block \cite{peebles2023scalable}. After the final block, the five slot features within each frame are concatenated and passed to stream-specific output heads to predict the clean interaction target $\hat{x}_0=D_{\theta}(x_t,c_D)$.
The frozen ManiVAE decoder maps each generated hand token back to a MANO pose, conditioned on the corresponding generated object-centered hand position and transformed object geometry.

We train DualFormer with clean-target prediction and component-wise supervision over the five semantic streams:

{\small
\begin{equation}
\begin{aligned}
&\mathcal L_{\mathrm{stream}}
=\norm{\hat O_{1:N}-O_{1:N}}_2^2\\
&+\sum_{h\in\{l,r\}}
\left(
\norm{\hat R_{1:N}^h-R_{1:N}^h}_2^2
+\norm{\hat z_{1:N}^h-z_{1:N}^h}_2^2
\right).
\end{aligned}
\end{equation}
}

Beyond rot6d reconstruction in the object stream, we supervise object orientation in $SO(3)$. Let $Q_i=\mathrm{Rot}(O_i^\beta)$ and $\hat Q_i=\mathrm{Rot}(\hat O_i^\beta)$; the geodesic loss is
{\small
\begin{equation}
\mathcal L_{\mathrm{rot}}
=\frac{1}{N}\sum_{i=1}^{N}
\arccos\!\left(
\frac{\operatorname{tr}(\hat Q_i^\top Q_i)-1}{2}
\right).
\end{equation}
}
The principal DualFormer objective is
{\small
\begin{equation}
\mathcal L_D
=\mathcal L_{\mathrm{stream}}
+\lambda_{\mathrm{rot}}\mathcal L_{\mathrm{rot}}.
\end{equation}
}
At inference, standard reverse sampling produces the global interaction passed to ProxiRefine.

\begin{figure}[tb]
    \centering
    \includegraphics[width=1\linewidth]{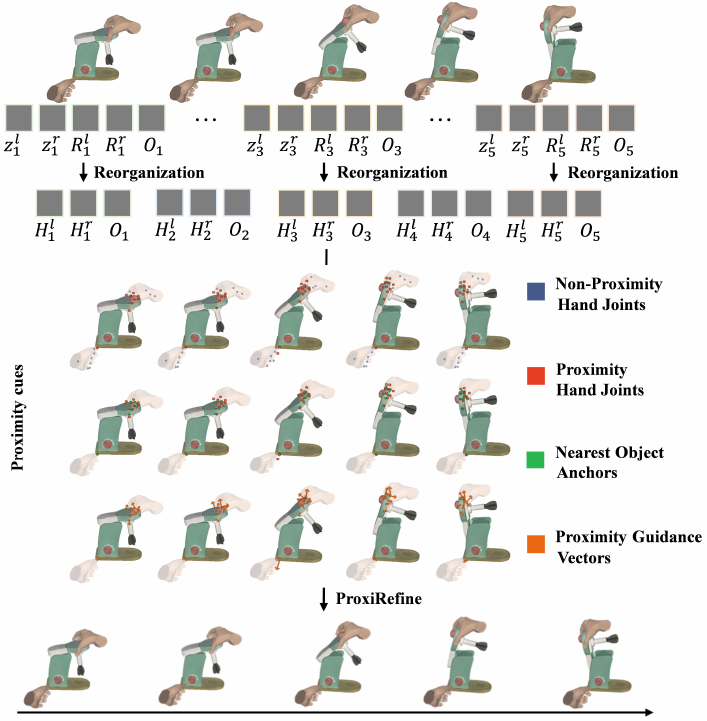}
    \caption{Object-fixed ProxiRefine.
    ProxiRefine derives joint-level proximity cues from the generated interaction and residually corrects bimanual position and pose while keeping the object trajectory fixed.}
    \label{fig:proxirefine}
\end{figure}


\subsection{Object-Fixed Hand Refinement}
\label{sec:proxirefine}

Decoded ManiVAE poses may still contain local contact errors \cite{grady2021contactopt,jiang2021hand,liu2023contactgen}. ProxiRefine fixes the DualFormer object trajectory $\hat O_{1:N}$ and predicts bimanual position and pose residuals from current hand--object proximity.

ProxiRefine operates in direct MANO pose space. Let $\hat{\mathcal O}_i$ denote the translation-free object geometry transformed according to the generated rotation and articulation at frame $i$. For each $h\in\{l,r\}$, we decode the generated hand-pose token as $\hat P_i^h=D_M(\hat z_i^h,\hat R_i^h,\hat{\mathcal O}_i)$, recover its global translation as $\hat T_i^h=\hat O_i^\alpha+\hat R_i^h$, and form $\hat H_i^h=\{\hat T_i^h,\hat P_i^h\}$.
ProxiRefine constructs joint-level proximity cues from the DualFormer result. Let $\mathcal{J}(\hat{H}_i^h)$ denote the MANO joints, and let $\Pi_{\hat{O}_i}(\cdot)$ return their nearest points on the generated object surface at frame $i$. We construct the proximity cue for hand $h$ as
{\small
\begin{equation}
C_i^h =
\left[
\mathcal{J}(\hat{H}_i^h)-\Pi_{\hat{O}_i}(\mathcal{J}(\hat{H}_i^h)),\;
d_i^h,\;
\exp(-\eta d_i^h)
\right],
\end{equation}
}
where $d_i^h$ is the joint-to-surface distance and $\eta$ controls the distance decay. The cue combines surface offsets, distances, and proximity weights, while a distance threshold defines the proximity mask. As visualized in Figure~\ref{fig:proxirefine}, the resulting proximity mask marks the selected joints in red and the remaining joints in dark blue. For each selected joint, the green anchor is its nearest point on the generated object surface, and the orange arrow represents the corresponding surface-to-joint offset vector. The refiner takes the generated hands, object sequence, and proximity cues as input:
{\small
\begin{equation}
H_{1:N}^{l,\mathrm{ref}},H_{1:N}^{r,\mathrm{ref}}
=
F_P(\hat{H}_{1:N}^l,\hat{H}_{1:N}^r,\hat{O}_{1:N},C_{1:N}^l,C_{1:N}^r).
\end{equation}
}
The refiner predicts residuals relative to its input hands, $H_i^{h,\mathrm{ref}}=\hat{H}_i^h+\Delta_i^h$ for $h\in\{l,r\}$.
Translation residuals are added directly, whereas pose residuals are applied to the 6D representation of each MANO joint rotation and then converted back to rotation matrices.
Temporal position encodings and hand-role embeddings couple the two hand streams. The residual heads are zero-initialized to preserve the input motion at the start of training.

The hand reconstruction term supervises relative placement and 6D MANO pose:
\begingroup
\fontsize{8pt}{9pt}\selectfont
\begin{equation}
\begin{aligned}
\mathcal{L}_{\mathrm{hand}}^{P}
={}&
\lambda_R^P
\left(
\norm{R_{1:N}^{l,\mathrm{ref}}-R_{1:N}^{l}}_2^{2}
+
\norm{R_{1:N}^{r,\mathrm{ref}}-R_{1:N}^{r}}_2^{2}
\right) \\
&+
\lambda_P^P
\left(
\norm{P_{1:N}^{l,\mathrm{ref}}-P_{1:N}^{l}}_2^{2}
+
\norm{P_{1:N}^{r,\mathrm{ref}}-P_{1:N}^{r}}_2^{2}
\right).
\end{aligned}
\end{equation}
\endgroup

Let $\mathcal{K}_i^h$ contain joints selected by the distance threshold, and let $|\mathcal{K}|=\sum_{h,i}|\mathcal{K}_i^h|$. We define the contact loss as
{\small
\begin{equation}
\begin{aligned}
\mathcal{L}_{\mathrm{contact}}
={}&
\frac{1}{\max(|\mathcal{K}|,1)}
\sum_{h\in\{l,r\}}\sum_i\sum_{j\in\mathcal{K}_i^h} \\
&\norm{
\mathcal{J}_{j}(H_i^{h,\mathrm{ref}})
-
\Pi_{\hat{O}_i}\!\left(\mathcal{J}_{j}(\hat{H}_i^h)\right)
}_1 .
\end{aligned}
\end{equation}
}

We set the loss to zero when $|\mathcal K|=0$. The surface targets are computed from the DualFormer output, preserving its contact assignment during refinement. The principal ProxiRefine objective is
{\small
\begin{equation}
\mathcal{L}_P
=\mathcal{L}_{\mathrm{hand}}^{P}
+\lambda_{\mathrm{contact}}\mathcal{L}_{\mathrm{contact}}.
\end{equation}
}

\begin{table*}[t]
\centering
{
\small
\setlength{\tabcolsep}{1.25mm}
\begin{tabular}{lllrrcrrrrrrr}
\toprule
Dataset & Setting & Method & Phy$\uparrow$ & C-F1$\uparrow$ & MPJPE$_{\mathrm{A}}$\,$\downarrow$ & OTE$\downarrow$ & ORE$\downarrow$ & SD$\uparrow$ & OD$\uparrow$ & C-Rate$\uparrow$ & C-Stab$\uparrow$ & NN-Rel$\downarrow$ \\
\midrule
\multirow{8}{*}{ARCTIC}
& \multirow{4}{*}{Bim.-Art.}
& Text2HOI & 0.830 & 0.698 & 13.901 & 0.405 & 12.670 & 4.540 & 8.338 & 0.619 & 0.955 & 0.227 \\
& & LatentHOI & 0.955 & 0.767 & 14.183 & 0.425 & 14.439 & 7.205 & 11.638 & 0.799 & 0.957 & 0.162 \\
& & MDM & 0.867 & 0.743 & 14.032 & 0.380 & 13.909 & \textbf{8.007} & \textbf{11.746} & 0.713 & 0.960 & 0.209 \\
& & DuetHOI (Ours) & \textbf{0.997} & \textbf{0.834} & \textbf{12.043} & \textbf{0.332} & \textbf{12.186} & 4.489 & 10.271 & \textbf{0.873} & \textbf{0.970} & \textbf{0.095} \\
\cmidrule(lr){2-13}
& \multirow{4}{*}{Bim.-Rig.}
& Text2HOI & 0.875 & 0.631 & 14.002 & 0.551 & \textbf{6.027} & 3.770 & 6.522 & 0.608 & \textbf{0.959} & 0.188 \\
& & LatentHOI & 0.845 & 0.694 & 14.329 & 0.545 & 9.438 & \textbf{6.859} & 10.243 & 0.642 & 0.954 & 0.178 \\
& & MDM & 0.810 & 0.643 & 12.879 & 0.422 & 6.993 & 6.370 & 10.562 & 0.572 & 0.957 & 0.193 \\
& & DuetHOI (Ours) & \textbf{1.000} & \textbf{0.746} & \textbf{9.614} & \textbf{0.357} & 7.301 & 4.047 & \textbf{10.722} & \textbf{0.705} & 0.956 &\textbf{ 0.118} \\
\midrule
\multirow{4}{*}{GRAB}
& \multirow{4}{*}{Bim.-Rig.}
& Text2HOI & 0.789 & 0.725 & 29.240 & 0.765 & \textbf{31.062} & 6.866 & 10.191 & 0.617 & 0.949 & 0.156 \\
& & LatentHOI & 0.752 & 0.644 & 30.966 & 1.247 & 87.927 & 5.008 & 6.375 & 0.527 & 0.938 & 0.162 \\
& & MDM & 0.658 & 0.580 & 31.029 & 0.786 & 31.238 & \textbf{18.048} & \textbf{21.801} & 0.444 & 0.981 & 0.186 \\
& & DuetHOI (Ours) & \textbf{0.975} & \textbf{0.775} & \textbf{28.806} & \textbf{0.748} & 48.124 & 9.318 & 15.858 & \textbf{0.657} & \textbf{0.986} & \textbf{0.088} \\
\bottomrule
\end{tabular}
}
\caption{Main bimanual results on ARCTIC and GRAB. Arrows indicate the preferred metric direction.}
\label{tab:bimanual_main_results}
\end{table*}

\begin{figure*}[tb]
    \centering
    \includegraphics[width=0.98\textwidth]{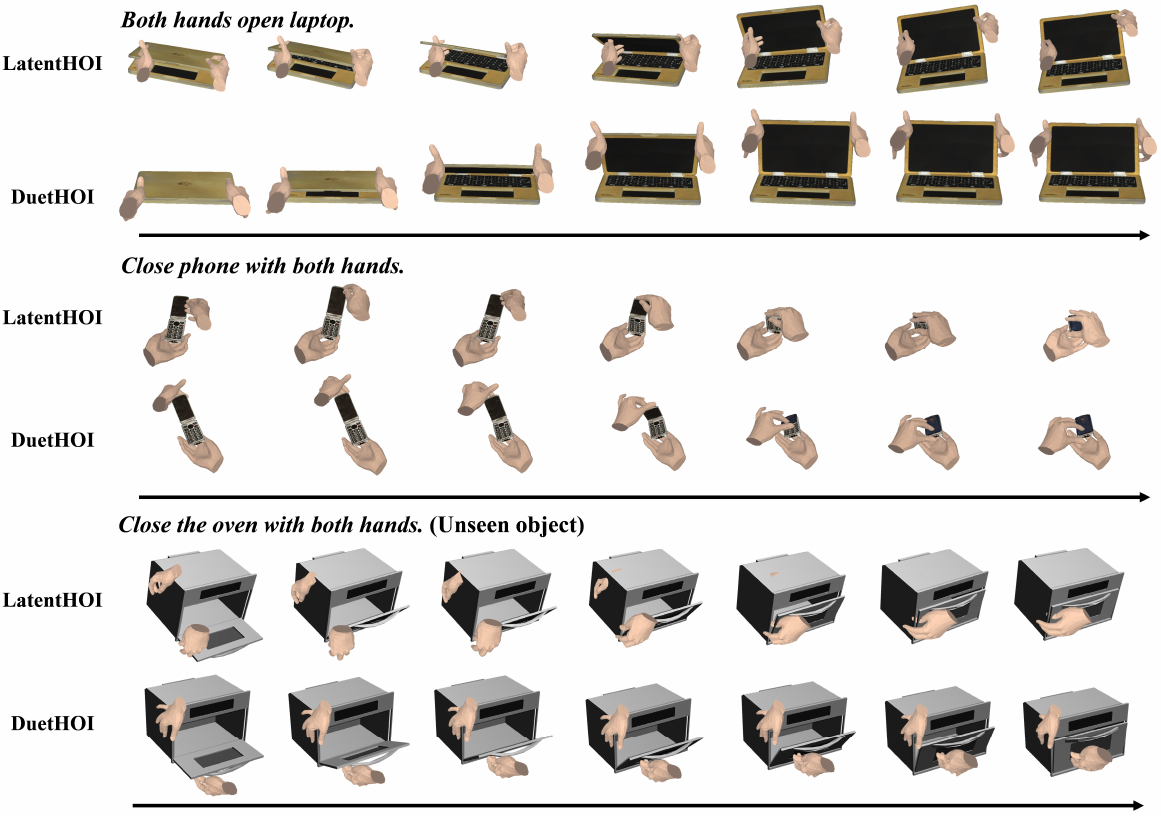}
    \caption{Qualitative results. DuetHOI preserves bimanual coordination during articulation, including for unseen ovens.}
    \label{fig:qualitative_results}
\end{figure*}

\section{Experiments}
\label{sec:experiments}

\subsection{Experimental Setup}
\label{sec:experimental_setup}

We evaluate on ARCTIC \cite{fan2023arctic} and GRAB \cite{taheri2020grab} using four modes: single-hand articulated (Single-Art.), single-hand rigid (Single-Rig.), bimanual articulated (Bim.-Art.), and bimanual rigid (Bim.-Rig.). ARCTIC covers all four modes, whereas GRAB is evaluated on rigid objects. Each model is jointly trained on all available modes and evaluated per mode; absent-hand streams are masked. Bim.-Art. is our primary setting, while the others test the unified representation across hand usage and articulation. We adapt Text2HOI \cite{cha2024text2hoi}, LatentHOI \cite{li2025latenthoi}, and MDM \cite{tevet2022human} to the same language-conditioned HOI task, providing the same raw task inputs, data splits, and evaluation protocol while preserving their core modeling principles. We report Phy, C-F1, MPJPE$_{\mathrm{A}}$, OTE, ORE, SD, OD, C-Rate, C-Stab, and NN-Rel for physical plausibility, contact, hand and object accuracy, diversity, stability, and relative hand--object consistency. Baseline adaptation and metric definitions are provided in the supplementary material.

\subsection{Main Results on Bimanual Interaction}
\label{sec:bimanual_main_results}

On ARCTIC Bim.-Art. in Table~\ref{tab:bimanual_main_results}, DuetHOI leads eight of ten metrics, with \textbf{0.997} Phy, \textbf{0.834} C-F1, and \textbf{0.095} NN-Rel. Against the strongest baselines, C-F1 improves by \textbf{0.067}, while MPJPE$_{\mathrm A}$ and NN-Rel decrease by \textbf{13.4\%} and \textbf{41.4\%}. C-Rate and C-Stab confirm persistent contact. These gains support separating articulation progress, object-centered bimanual placement, and local hand pose before object-fixed contact refinement. Rigid-interaction gains persist: on ARCTIC Bim.-Rig., DuetHOI reaches \textbf{0.746} C-F1 and reduces MPJPE$_{\mathrm A}$ from 12.879 to \textbf{9.614} by \textbf{25.4\%}; on GRAB, it reaches \textbf{0.775} C-F1 and lowers NN-Rel from 0.156 to \textbf{0.088} by \textbf{43.6\%}. Baselines retain advantages in rotation accuracy or diversity. Figure~\ref{fig:qualitative_results} shows coherent bimanual manipulation as articulation evolves, including the unseen oven category.

\begin{table}[t]
\centering
{
\small
\begin{tabular}{@{}lrrcrr@{}}
\toprule
Method & Phy$\uparrow$ & C-F1$\uparrow$ & MPJPE$_{\mathrm{A}}$\,$\downarrow$ & OTE$\downarrow$ & ORE$\downarrow$ \\
\midrule
\multicolumn{6}{@{}l}{\textit{ARCTIC, Single-Art.}} \\
Text2HOI & 0.664 & 0.518 & 12.535 & 0.504 & 11.587 \\
LatentHOI & 0.757 & 0.569 & 13.249 & 0.488 & 13.550 \\
MDM & 0.702 & 0.571 & 11.647 & 0.425 & 11.663 \\
DuetHOI (Ours) & \textbf{0.992} & \textbf{0.660} & \textbf{9.573} & \textbf{0.361} & \textbf{11.411} \\
\midrule
\multicolumn{6}{@{}l}{\textit{ARCTIC, Single-Rig.}} \\
Text2HOI & 0.750 & 0.471 & 14.403 & 0.497 & 10.834 \\
LatentHOI & 0.689 & 0.559 & 15.901 & 0.602 & 12.106 \\
MDM & 0.700 & 0.515 & 14.211 & 0.520 & 11.461 \\
DuetHOI (Ours) & \textbf{0.947} & \textbf{0.633} & \textbf{11.875} & \textbf{0.481} & \textbf{10.454} \\
\midrule
\multicolumn{6}{@{}l}{\textit{GRAB, Single-Rig.}} \\
Text2HOI & 0.646 & 0.580 & 14.594 & 1.178 & 31.749 \\
LatentHOI & 0.678 & 0.608 & 15.253 & 1.393 & 58.235 \\
MDM & 0.699 & 0.599 & 14.699 & 1.078 & \textbf{31.560} \\
DuetHOI (Ours) & \textbf{0.981} & \textbf{0.783} & \textbf{14.286} & \textbf{1.043} & 46.262 \\
\bottomrule
\end{tabular}
}
\caption{Single-hand results of the jointly trained model; full metrics are in the supplement.}
\label{tab:single_hand_generalization}
\end{table}

\subsection{Performance on Single-Hand Interaction}
\label{sec:single_hand_generalization}

In Table~\ref{tab:single_hand_generalization}, with one hand stream masked, DuetHOI achieves \textbf{0.992} Phy, \textbf{0.660} C-F1, and \textbf{9.573} MPJPE$_{\mathrm A}$ on ARCTIC Single-Art., outperforming all baselines. It leads all five metrics on ARCTIC Single-Rig., including \textbf{0.947} Phy and \textbf{0.633} C-F1. On GRAB, it raises C-F1 from 0.608 to \textbf{0.783}, although MDM obtains a lower ORE. This robustness follows from the semantic-stream design: masking the absent hand preserves object and active-hand streams, while ManiVAE provides a common compact pose prior for either hand.

\subsection{Articulation-Aware Interaction Generation}
\label{sec:articulation_aware_generation}

Before ProxiRefine, full DualFormer reaches \textbf{0.969} Phy and \textbf{0.786} C-F1 on Bim.-Art. in Table~\ref{tab:dualformer_ablation}. Without ContactVAE, Phy/C-F1 fall to 0.836/0.679 on Bim.-Art. and 0.765/0.683 on Bim.-Rig., confirming its surface prior anchors placement across articulation types. Without ArtPlanner, articulated MPJPE$_{\mathrm A}$ increases from \textbf{12.243} to 13.272 because its trajectory reference aligns hand motion with evolving geometry. Restricting output heads to their own slots lowers C-F1 to 0.776/0.670 and raises MPJPE$_{\mathrm A}$ to 12.628/11.060, exposing weaker cross-stream coordination. The larger rigid MPJPE increase without ManiVAE highlights compact pose tokens, especially without an articulation reference. Flattened attention raises rigid MPJPE$_{\mathrm A}$ to 11.745, supporting factorization. Overall, results support cross-stream context, compact pose tokens, and factorized attention.

\begin{table}[t]
\centering
{\small
\setlength{\tabcolsep}{1mm}
\begin{tabular}{@{}lrrcrr@{}}
\toprule
Variant
& Phy$\uparrow$
& C-F1$\uparrow$
& MPJPE$_{\mathrm{A}}$\,$\downarrow$
& OTE$\downarrow$
& ORE$\downarrow$ \\
\midrule
\multicolumn{6}{@{}l}{\textit{Bim.-Art.}} \\
w/o ManiVAE
& 0.951 & 0.775 & 12.458 & 0.341 & 12.191 \\
w/o ContactVAE
& 0.836 & 0.679 & 13.690 & 0.346 & 12.212 \\
w/o ArtPlanner
& 0.962 & 0.779 & 13.272 & 0.342 & 12.202 \\
w/ slot context
& 0.952 & 0.776 & 12.628 & 0.333 & \textbf{11.872} \\
Flattened atten.
& 0.968 & 0.784 & 12.429 & 0.335 & 12.995 \\
Full
& \textbf{0.969} & \textbf{0.786} & \textbf{12.243}
& \textbf{0.332} & 12.186 \\
\midrule
\multicolumn{6}{@{}l}{\textit{Bim.-Rig.}} \\
w/o ManiVAE
& 0.833 & 0.700 & 10.512 & 0.367 & 7.373 \\
w/o ContactVAE
& 0.765 & 0.683 & 9.700 & 0.376 & 7.447 \\
w/ slot context
& 0.828 & 0.670 & 11.060 & 0.360 & 8.369 \\
Flattened atten.
& 0.839 & \textbf{0.713} & 11.745 & 0.375 & 7.356 \\
Full
& \textbf{0.842} & \textbf{0.713} & \textbf{9.636}
& \textbf{0.357} & \textbf{7.301} \\
\bottomrule
\end{tabular}
}
\caption{DualFormer ablation on ARCTIC before ProxiRefine. Rigid objects omit ArtPlanner.}
\label{tab:dualformer_ablation}
\end{table}

\begin{table}[t]
\centering
{\small
\setlength{\tabcolsep}{1mm}
\begin{tabular}{@{}lrcrrr@{}}
\toprule
Variant
& C-F1$\uparrow$
& MPJPE$_{\mathrm{A}}$\,$\downarrow$
& C-Rate$\uparrow$
& C-Stab$\uparrow$
& NN-Rel$\downarrow$ \\
\midrule
\multicolumn{6}{@{}l}{\textit{Bim.-Art.}} \\
w/o $\mathcal L_{\mathrm{contact}}$
& 0.797 & 12.119 & 0.829 & 0.962 & 0.113 \\
w/o $C_i^h$
& 0.827 & 12.065 & 0.859 & 0.968 & 0.103 \\
Full
& \textbf{0.834} & \textbf{12.043} & \textbf{0.873}
& \textbf{0.970} & \textbf{0.095} \\
\midrule
\multicolumn{6}{@{}l}{\textit{Bim.-Rig.}} \\
w/o $\mathcal L_{\mathrm{contact}}$
& 0.689 & 9.622 & 0.632 & 0.950 & \textbf{0.112} \\
w/o $C_i^h$
& 0.730 & 9.652 & 0.688 & 0.954 & 0.119 \\
Full
& \textbf{0.746} & \textbf{9.614} & \textbf{0.705}
& \textbf{0.956} & 0.118 \\
\bottomrule
\end{tabular}
}
\caption{Ablation of the contact loss and joint-level proximity cue in object-fixed ProxiRefine on ARCTIC.}
\label{tab:proxirefine_ablation}
\end{table}

\subsection{Object-Fixed Proximity Refinement}
\label{sec:object_fixed_refinement}

Table~\ref{tab:proxirefine_ablation} shows that ProxiRefine raises C-F1 from 0.786/0.713 to \textbf{0.834/0.746} on Bim.-Art./Bim.-Rig., while MPJPE$_{\mathrm A}$ changes only from 12.243/9.636 to \textbf{12.043/9.614}. Thus, refinement improves contact without rewriting global motion. Moreover, C-Stab reaches \textbf{0.970/0.956}, while the fixed object trajectory attributes these gains to hand-geometry correction rather than altered manipulation progress. Removing $\mathcal L_{\mathrm{contact}}$ lowers C-F1 to 0.797/0.689 and C-Rate to 0.829/0.632, making direct contact supervision the primary refinement signal. Removing $C_i^h$ lowers C-F1 to 0.827/0.730 and C-Rate to 0.859/0.688; this supports the cue's localization role. Its slight rigid NN-Rel advantage (0.112 vs.\ \textbf{0.118}) comes at substantially lower C-F1 and C-Rate.

\section{Conclusion}
\label{sec:conclusion}

We presented DuetHOI for language-conditioned bimanual articulated-object interaction generation. It decomposes motion into object state, object-centered hand placement, and compact local hand articulation. ContactVAE and ArtPlanner provide spatial contact intent and a trajectory-level articulation reference, while ManiVAE and DualFormer generate coordinated global interaction. ProxiRefine corrects local hand geometry while fixing the object trajectory. Experiments on ARCTIC and GRAB show consistent gains in physical plausibility, contact quality, hand accuracy, and relative hand--object consistency, especially for bimanual articulated interactions. These results support structured generation with object-fixed local refinement.

\paragraph{Limitations.}
DuetHOI represents articulation with one scalar and generates one instruction-conditioned interaction at a time. Multi-joint objects require vector-valued planning and constraints among movable parts, whereas multi-task sequences require consistent contacts, hand roles, and object states across stages. The current model captures neither coupled joints nor inter-task transitions. Future work will study hierarchical multi-joint planning, coupled-part constraints, and compositional manipulation over longer task horizons.

\bibliography{aaai2027}

@inproceedings{li2025maniptrans,
  title={Maniptrans: Efficient dexterous bimanual manipulation transfer via residual learning},
  author={Li, Kailin and Li, Puhao and Liu, Tengyu and Li, Yuyang and Huang, Siyuan},
  booktitle={Proceedings of the Computer Vision and Pattern Recognition Conference},
  pages={6991--7003},
  year={2025}
}

@inproceedings{zhang2025mikudance,
  title={Mikudance: Animating character art with mixed motion dynamics},
  author={Zhang, Jiaxu and Zeng, Xianfang and Chen, Xin and Zuo, Wei and Yu, Gang and Tu, Zhigang},
  booktitle={Proceedings of the IEEE/CVF International Conference on Computer Vision},
  pages={19689--19699},
  year={2025}
}

@inproceedings{yang2022oakink,
  title={Oakink: A large-scale knowledge repository for understanding hand-object interaction},
  author={Yang, Lixin and Li, Kailin and Zhan, Xinyu and Wu, Fei and Xu, Anran and Liu, Liu and Lu, Cewu},
  booktitle={Proceedings of the IEEE/CVF conference on computer vision and pattern recognition},
  pages={20953--20962},
  year={2022}
}

@inproceedings{zhang2024artigrasp,
  title={Artigrasp: Physically plausible synthesis of bi-manual dexterous grasping and articulation},
  author={Zhang, Hui and Christen, Sammy and Fan, Zicong and Zheng, Luocheng and Hwangbo, Jemin and Song, Jie and Hilliges, Otmar},
  booktitle={2024 International Conference on 3D Vision (3DV)},
  pages={235--246},
  year={2024},
  organization={IEEE}
}

@inproceedings{zhang2024graspxl,
  title={Graspxl: Generating grasping motions for diverse objects at scale},
  author={Zhang, Hui and Christen, Sammy and Fan, Zicong and Hilliges, Otmar and Song, Jie},
  booktitle={European Conference on Computer Vision},
  pages={386--403},
  year={2024},
  organization={Springer}
}

@inproceedings{xu2023unidexgrasp,
  title={Unidexgrasp: Universal robotic dexterous grasping via learning diverse proposal generation and goal-conditioned policy},
  author={Xu, Yinzhen and Wan, Weikang and Zhang, Jialiang and Liu, Haoran and Shan, Zikang and Shen, Hao and Wang, Ruicheng and Geng, Haoran and Weng, Yijia and Chen, Jiayi and others},
  booktitle={Proceedings of the IEEE/CVF Conference on Computer Vision and Pattern Recognition},
  pages={4737--4746},
  year={2023}
}

@inproceedings{christen2022d,
  title={D-grasp: Physically plausible dynamic grasp synthesis for hand-object interactions},
  author={Christen, Sammy and Kocabas, Muhammed and Aksan, Emre and Hwangbo, Jemin and Song, Jie and Hilliges, Otmar},
  booktitle={Proceedings of the IEEE/CVF Conference on Computer Vision and Pattern Recognition},
  pages={20577--20586},
  year={2022}
}

@inproceedings{li2024semgrasp,
  title={Semgrasp: Semantic grasp generation via language aligned discretization},
  author={Li, Kailin and Wang, Jingbo and Yang, Lixin and Lu, Cewu and Dai, Bo},
  booktitle={European Conference on Computer Vision},
  pages={109--127},
  year={2024},
  organization={Springer}
}

@inproceedings{jiang2021hand,
  title={Hand-object contact consistency reasoning for human grasps generation},
  author={Jiang, Hanwen and Liu, Shaowei and Wang, Jiashun and Wang, Xiaolong},
  booktitle={Proceedings of the IEEE/CVF international conference on computer vision},
  pages={11107--11116},
  year={2021}
}

@article{liu2021synthesizing,
  title={Synthesizing diverse and physically stable grasps with arbitrary hand structures using differentiable force closure estimator},
  author={Liu, Tengyu and Liu, Zeyu and Jiao, Ziyuan and Zhu, Yixin and Zhu, Song-Chun},
  journal={IEEE Robotics and Automation Letters},
  volume={7},
  number={1},
  pages={470--477},
  year={2021},
  publisher={IEEE}
}

@inproceedings{taheri2020grab,
  title={GRAB: A dataset of whole-body human grasping of objects},
  author={Taheri, Omid and Ghorbani, Nima and Black, Michael J and Tzionas, Dimitrios},
  booktitle={European conference on computer vision},
  pages={581--600},
  year={2020},
  organization={Springer}
}

@inproceedings{taheri2022goal,
  title={GOAL: Generating 4D whole-body motion for hand-object grasping},
  author={Taheri, Omid and Choutas, Vasileios and Black, Michael J and Tzionas, Dimitrios},
  booktitle={Proceedings of the IEEE/CVF Conference on Computer Vision and Pattern Recognition},
  pages={13263--13273},
  year={2022}
}

@inproceedings{corona2020ganhand,
  title={Ganhand: Predicting human grasp affordances in multi-object scenes},
  author={Corona, Enric and Pumarola, Albert and Alenya, Guillem and Moreno-Noguer, Francesc and Rogez, Gr{\'e}gory},
  booktitle={Proceedings of the IEEE/CVF conference on computer vision and pattern recognition},
  pages={5031--5041},
  year={2020}
}

@inproceedings{turpin2022grasp,
  title={Grasp’d: Differentiable contact-rich grasp synthesis for multi-fingered hands},
  author={Turpin, Dylan and Wang, Liquan and Heiden, Eric and Chen, Yun-Chun and Macklin, Miles and Tsogkas, Stavros and Dickinson, Sven and Garg, Animesh},
  booktitle={European Conference on Computer Vision},
  pages={201--221},
  year={2022},
  organization={Springer}
}

@inproceedings{wu2022saga,
  title={Saga: Stochastic whole-body grasping with contact},
  author={Wu, Yan and Wang, Jiahao and Zhang, Yan and Zhang, Siwei and Hilliges, Otmar and Yu, Fisher and Tang, Siyu},
  booktitle={European Conference on Computer Vision},
  pages={257--274},
  year={2022},
  organization={Springer}
}

@article{tevet2022human,
  title={Human motion diffusion model},
  author={Tevet, Guy and Raab, Sigal and Gordon, Brian and Shafir, Yonatan and Cohen-Or, Daniel and Bermano, Amit H},
  journal={arXiv preprint arXiv:2209.14916},
  year={2022}
}

@inproceedings{zhou2024emdm,
  title={Emdm: Efficient motion diffusion model for fast and high-quality motion generation},
  author={Zhou, Wenyang and Dou, Zhiyang and Cao, Zeyu and Liao, Zhouyingcheng and Wang, Jingbo and Wang, Wenjia and Liu, Yuan and Komura, Taku and Wang, Wenping and Liu, Lingjie},
  booktitle={European Conference on Computer Vision},
  pages={18--38},
  year={2024},
  organization={Springer}
}

@article{romero2022embodied,
  title={Embodied hands: Modeling and capturing hands and bodies together},
  author={Romero, Javier and Tzionas, Dimitrios and Black, Michael J},
  journal={arXiv preprint arXiv:2201.02610},
  year={2022}
}

@inproceedings{zhou2019continuity,
  title={On the continuity of rotation representations in neural networks},
  author={Zhou, Yi and Barnes, Connelly and Lu, Jingwan and Yang, Jimei and Li, Hao},
  booktitle={Proceedings of the IEEE/CVF conference on computer vision and pattern recognition},
  pages={5745--5753},
  year={2019}
}

@inproceedings{milletari2016v,
  title={V-net: Fully convolutional neural networks for volumetric medical image segmentation},
  author={Milletari, Fausto and Navab, Nassir and Ahmadi, Seyed-Ahmad},
  booktitle={2016 fourth international conference on 3D vision (3DV)},
  pages={565--571},
  year={2016},
  organization={Ieee}
}

@inproceedings{peebles2023scalable,
  title={Scalable diffusion models with transformers},
  author={Peebles, William and Xie, Saining},
  booktitle={Proceedings of the IEEE/CVF international conference on computer vision},
  pages={4195--4205},
  year={2023}
}

@inproceedings{radford2021learning,
  title={Learning transferable visual models from natural language supervision},
  author={Radford, Alec and Kim, Jong Wook and Hallacy, Chris and Ramesh, Aditya and Goh, Gabriel and Agarwal, Sandhini and Sastry, Girish and Askell, Amanda and Mishkin, Pamela and Clark, Jack and others},
  booktitle={International conference on machine learning},
  pages={8748--8763},
  year={2021},
  organization={PmLR}
}

@article{ho2020denoising,
  title={Denoising diffusion probabilistic models},
  author={Ho, Jonathan and Jain, Ajay and Abbeel, Pieter},
  journal={Advances in neural information processing systems},
  volume={33},
  pages={6840--6851},
  year={2020}
}

@article{yang2024learning,
  title={Learning a contact potential field for modeling the hand-object interaction},
  author={Yang, Lixin and Zhan, Xinyu and Li, Kailin and Xu, Wenqiang and Zhang, Junming and Li, Jiefeng and Lu, Cewu},
  journal={IEEE transactions on pattern analysis and machine intelligence},
  volume={46},
  number={8},
  pages={5645--5662},
  year={2024},
  publisher={IEEE}
}

@inproceedings{cha2024text2hoi,
  title={Text2hoi: Text-guided 3d motion generation for hand-object interaction},
  author={Cha, Junuk and Kim, Jihyeon and Yoon, Jae Shin and Baek, Seungryul},
  booktitle={Proceedings of the IEEE/CVF Conference on Computer Vision and Pattern Recognition},
  pages={1577--1585},
  year={2024}
}

@inproceedings{li2025latenthoi,
  title={LatentHOI: On the Generalizable Hand Object Motion Generation with Latent Hand Diffusion.},
  author={Li, Muchen and Christen, Sammy and Wan, Chengde and Cai, Yujun and Liao, Renjie and Sigal, Leonid and Ma, Shugao},
  booktitle={Proceedings of the Computer Vision and Pattern Recognition Conference},
  pages={17416--17425},
  year={2025}
}

@inproceedings{ye2023affordance,
  title={Affordance diffusion: Synthesizing hand-object interactions},
  author={Ye, Yufei and Li, Xueting and Gupta, Abhinav and De Mello, Shalini and Birchfield, Stan and Song, Jiaming and Tulsiani, Shubham and Liu, Sifei},
  booktitle={Proceedings of the IEEE/CVF Conference on Computer Vision and Pattern Recognition},
  pages={22479--22489},
  year={2023}
}

@inproceedings{huang2025hoigpt,
  title={HOIGPT: Learning Long-Sequence Hand-Object Interaction with Language Models},
  author={Huang, Mingzhen and Chu, Fu-Jen and Tekin, Bugra and Liang, Kevin J and Ma, Haoyu and Wang, Weiyao and Chen, Xingyu and Gleize, Pierre and Xue, Hongfei and Lyu, Siwei and others},
  booktitle={Proceedings of the Computer Vision and Pattern Recognition Conference},
  pages={7136--7146},
  year={2025}
}

@inproceedings{zheng2023cams,
  title={Cams: Canonicalized manipulation spaces for category-level functional hand-object manipulation synthesis},
  author={Zheng, Juntian and Zheng, Qingyuan and Fang, Lixing and Liu, Yun and Yi, Li},
  booktitle={Proceedings of the IEEE/CVF Conference on Computer Vision and Pattern Recognition},
  pages={585--594},
  year={2023}
}

@inproceedings{fan2023arctic,
  title={ARCTIC: A dataset for dexterous bimanual hand-object manipulation},
  author={Fan, Zicong and Taheri, Omid and Tzionas, Dimitrios and Kocabas, Muhammed and Kaufmann, Manuel and Black, Michael J and Hilliges, Otmar},
  booktitle={Proceedings of the IEEE/CVF conference on computer vision and pattern recognition},
  pages={12943--12954},
  year={2023}
}

@inproceedings{ludwig2025efficient,
  title={Efficient 2D to full 3D human pose uplifting including joint rotations},
  author={Ludwig, Katja and Oksymets, Yuliia and Sch{\"o}n, Robin and Kienzle, Daniel and Lienhart, Rainer},
  booktitle={Proceedings of the Computer Vision and Pattern Recognition Conference},
  pages={5852--5861},
  year={2025}
}

@inproceedings{rempe2021humor,
  title={Humor: 3d human motion model for robust pose estimation},
  author={Rempe, Davis and Birdal, Tolga and Hertzmann, Aaron and Yang, Jimei and Sridhar, Srinath and Guibas, Leonidas J},
  booktitle={Proceedings of the IEEE/CVF international conference on computer vision},
  pages={11488--11499},
  year={2021}
}

@inproceedings{petrovich2021action,
  title={Action-conditioned 3d human motion synthesis with transformer vae},
  author={Petrovich, Mathis and Black, Michael J and Varol, G{\"u}l},
  booktitle={Proceedings of the IEEE/CVF international conference on computer vision},
  pages={10985--10995},
  year={2021}
}

@inproceedings{petrovich2022temos,
  title={Temos: Generating diverse human motions from textual descriptions},
  author={Petrovich, Mathis and Black, Michael J and Varol, G{\"u}l},
  booktitle={European conference on computer vision},
  pages={480--497},
  year={2022},
  organization={Springer}
}

@inproceedings{guo2022generating,
  title={Generating diverse and natural 3d human motions from text},
  author={Guo, Chuan and Zou, Shihao and Zuo, Xinxin and Wang, Sen and Ji, Wei and Li, Xingyu and Cheng, Li},
  booktitle={Proceedings of the IEEE/CVF conference on computer vision and pattern recognition},
  pages={5152--5161},
  year={2022}
}

@inproceedings{tevet2022motionclip,
  title={Motionclip: Exposing human motion generation to clip space},
  author={Tevet, Guy and Gordon, Brian and Hertz, Amir and Bermano, Amit H and Cohen-Or, Daniel},
  booktitle={European Conference on Computer Vision},
  pages={358--374},
  year={2022},
  organization={Springer}
}

@inproceedings{zhang2023generating,
  title={Generating human motion from textual descriptions with discrete representations},
  author={Zhang, Jianrong and Zhang, Yangsong and Cun, Xiaodong and Zhang, Yong and Zhao, Hongwei and Lu, Hongtao and Shen, Xi and Shan, Ying},
  booktitle={Proceedings of the IEEE/CVF conference on computer vision and pattern recognition},
  pages={14730--14740},
  year={2023}
}

@inproceedings{guo2024momask,
  title={Momask: Generative masked modeling of 3d human motions},
  author={Guo, Chuan and Mu, Yuxuan and Javed, Muhammad Gohar and Wang, Sen and Cheng, Li},
  booktitle={Proceedings of the IEEE/CVF Conference on Computer Vision and Pattern Recognition},
  pages={1900--1910},
  year={2024}
}

@inproceedings{moon2019posefix,
  title={Posefix: Model-agnostic general human pose refinement network},
  author={Moon, Gyeongsik and Chang, Ju Yong and Lee, Kyoung Mu},
  booktitle={Proceedings of the IEEE/CVF Conference on Computer Vision and Pattern Recognition},
  pages={7773--7781},
  year={2019}
}

@inproceedings{grady2021contactopt,
  title={Contactopt: Optimizing contact to improve grasps},
  author={Grady, Patrick and Tang, Chengcheng and Twigg, Christopher D and Vo, Minh and Brahmbhatt, Samarth and Kemp, Charles C},
  booktitle={Proceedings of the IEEE/CVF Conference on Computer Vision and Pattern Recognition},
  pages={1471--1481},
  year={2021}
}

@inproceedings{liu2023contactgen,
  title={Contactgen: Generative contact modeling for grasp generation},
  author={Liu, Shaowei and Zhou, Yang and Yang, Jimei and Gupta, Saurabh and Wang, Shenlong},
  booktitle={Proceedings of the IEEE/CVF International Conference on Computer Vision},
  pages={20609--20620},
  year={2023}
}

@inproceedings{arnab2021vivit,
  title={Vivit: A video vision transformer},
  author={Arnab, Anurag and Dehghani, Mostafa and Heigold, Georg and Sun, Chen and Lu{\v{c}}i{\'c}, Mario and Schmid, Cordelia},
  booktitle={Proceedings of the IEEE/CVF international conference on computer vision},
  pages={6836--6846},
  year={2021}
}

@inproceedings{li2025object,
  title={Object-centric prompt-driven vision-language-action model for robotic manipulation},
  author={Li, Xiaoqi and Xu, Jingyun and Zhang, Mingxu and Liu, Jiaming and Shen, Yan and Ponomarenko, Iaroslav and Xu, Jiahui and Heng, Liang and Huang, Siyuan and Zhang, Shanghang and others},
  booktitle={Proceedings of the IEEE/CVF Conference on Computer Vision and Pattern Recognition},
  pages={27638--27648},
  year={2025}
}

@article{qi2017pointnet++,
  title={Pointnet++: Deep hierarchical feature learning on point sets in a metric space},
  author={Qi, Charles Ruizhongtai and Yi, Li and Su, Hao and Guibas, Leonidas J},
  journal={Advances in neural information processing systems},
  volume={30},
  year={2017}
}


\end{document}